\documentclass[runningheads]{llncs}
\usepackage{graphicx}

\usepackage{tikz}
\usepackage{comment}
\usepackage{amsmath,amssymb} 
\usepackage{color}

\usepackage[accsupp]{axessibility}  

\usepackage{subfig}
\usepackage[linesnumbered,ruled,vlined]{algorithm2e}

\usepackage{newfloat}
\usepackage{listings}
\usepackage{subfloat}
\usepackage{multicol}
\usepackage{float}
\usepackage{bm}
\usepackage{booktabs}
\usepackage{marvosym}

\usepackage[pagebackref,breaklinks,colorlinks]{hyperref}
\usepackage{microtype}

\usepackage[capitalize]{cleveref}
\crefname{section}{Sec.}{Secs.}
\Crefname{section}{Section}{Sections}
\Crefname{table}{Table}{Tables}
\crefname{table}{Tab.}{Tabs.}

\def\Pmore#1{{\textcolor[rgb]{1,0.2,0.2}{$\uparrow$#1}}}
\def\Pless#1{{\textcolor[rgb]{0,0.6,0.41}{$\downarrow$#1}}}
\def\UT#1#2{$\text{#1}^\text{#2}$}

\begin{document}
\pagestyle{headings}
\mainmatter
\def\ECCVSubNumber{2179}  

\title{Adversarially-Aware Robust Object Detector \\ \vspace{-2ex}} 

\titlerunning{Adversarially-Aware Robust Object Detector}
%
\author{
Ziyi Dong \and
Pengxu Wei\thanks{Corresponding Author} \and
Liang Lin}
\authorrunning{Ziyi Dong, Pengxu Wei, Liang Lin}
%
\institute{Sun Yat-Sen University, Guangzhou, China \\ \email{dongzy6@mail2.sysu.edu.cn, weipx3@mail.sysu.edu.cn, linliang@ieee.org}}
\maketitle

\vspace{-13mm}
\hspace{-20pt}
\begin{minipage}{1.0\textwidth}
    \begin{figure}[H]
        \centering
        \hspace{-12pt}
        \subfloat[Clean images]{
        \begin{minipage}[t]{0.332\linewidth} \label{fig:architecture1}
        \centering
        \includegraphics[width=1\textwidth]{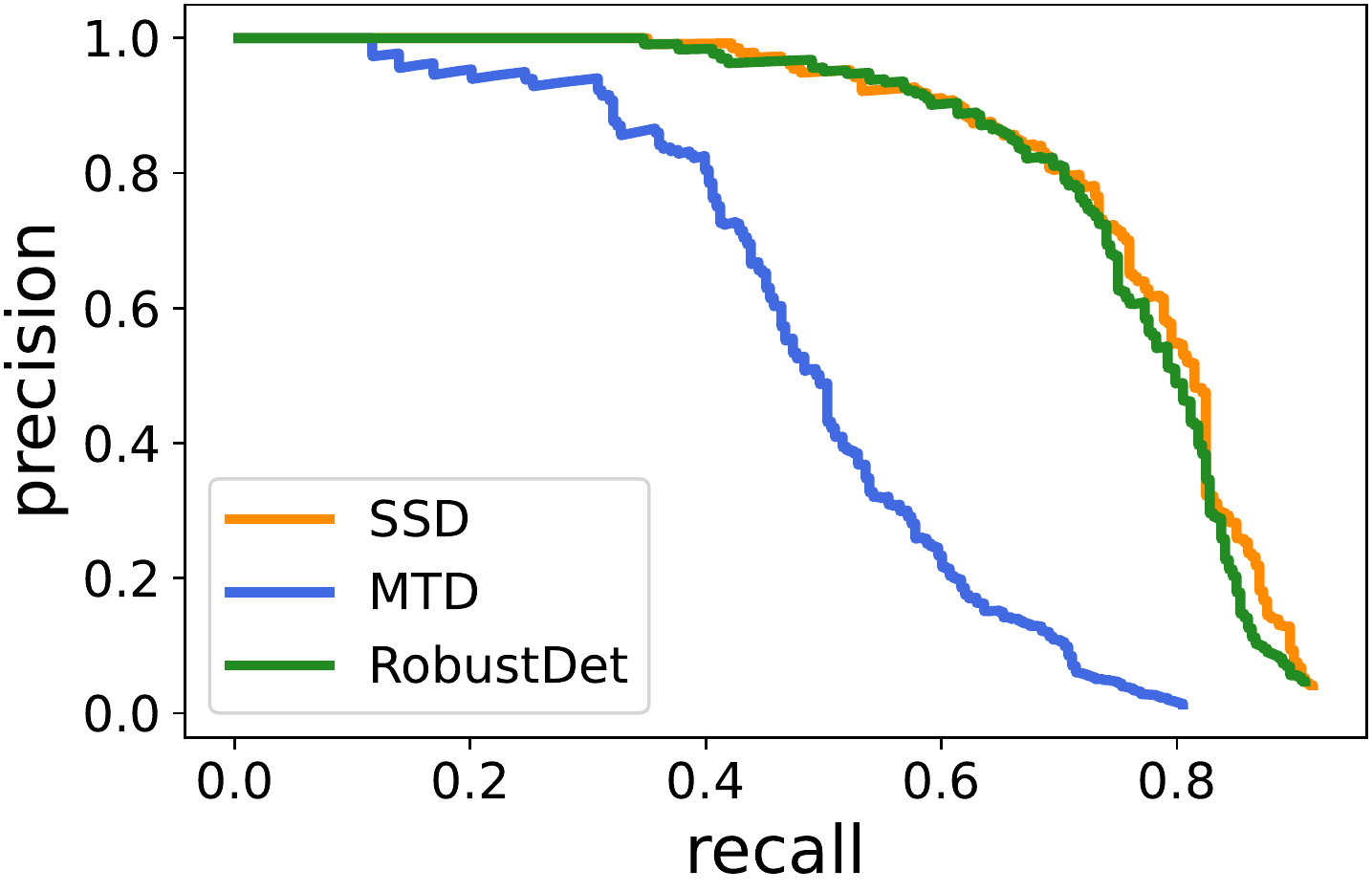}
        \end{minipage}%
        }%
        \subfloat[Adversarial images($A_{cls}$)]{
        \begin{minipage}[t]{0.332\linewidth}\label{fig:architecture2}
        \centering
        \includegraphics[width=1\textwidth]{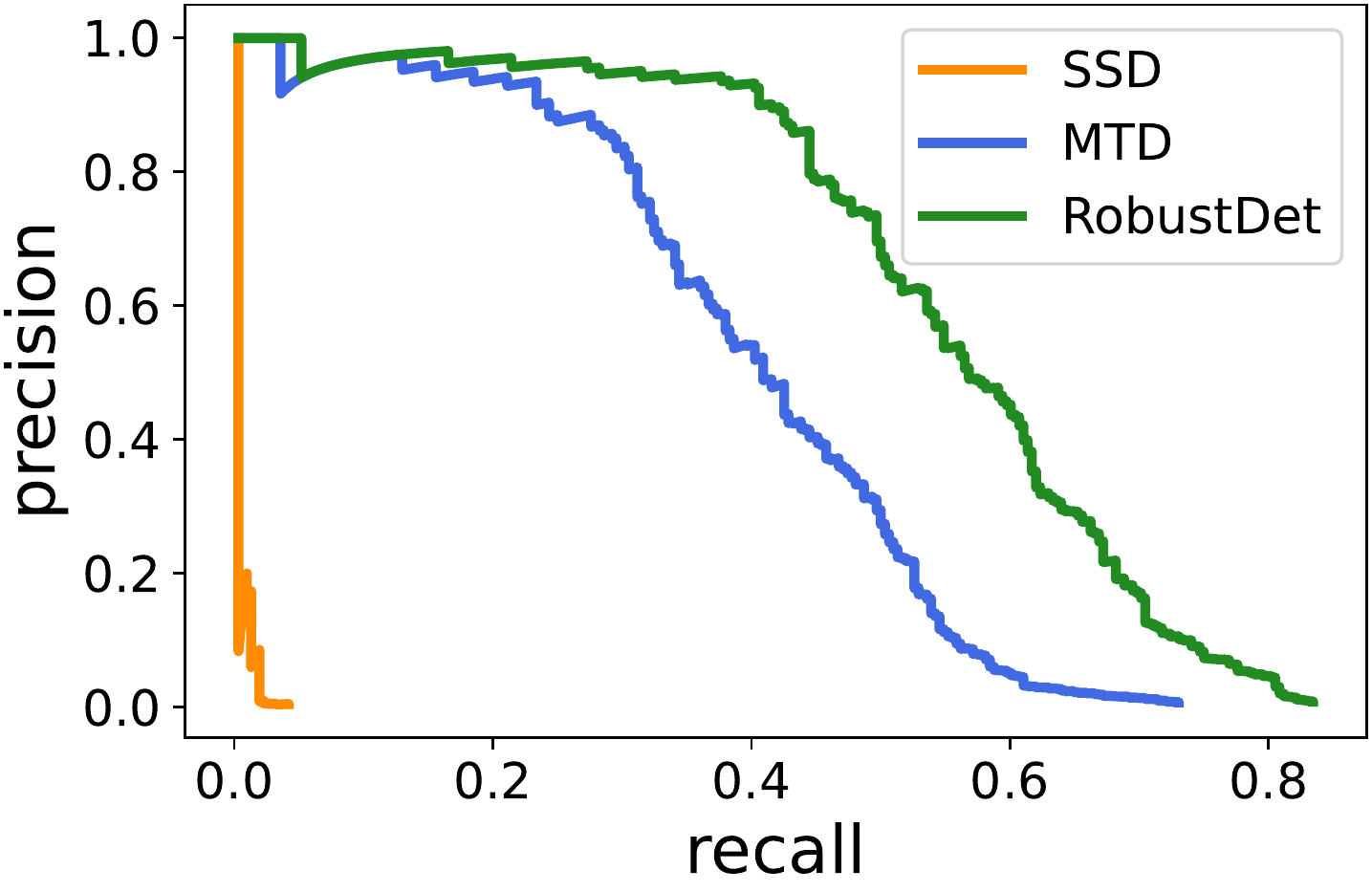}
        \end{minipage}%
        }%
        \subfloat[Adversarial images($A_{loc}$)]{
        \begin{minipage}[t]{0.332\linewidth} \label{fig:architecture3}
        \centering
        \includegraphics[width=1\textwidth]{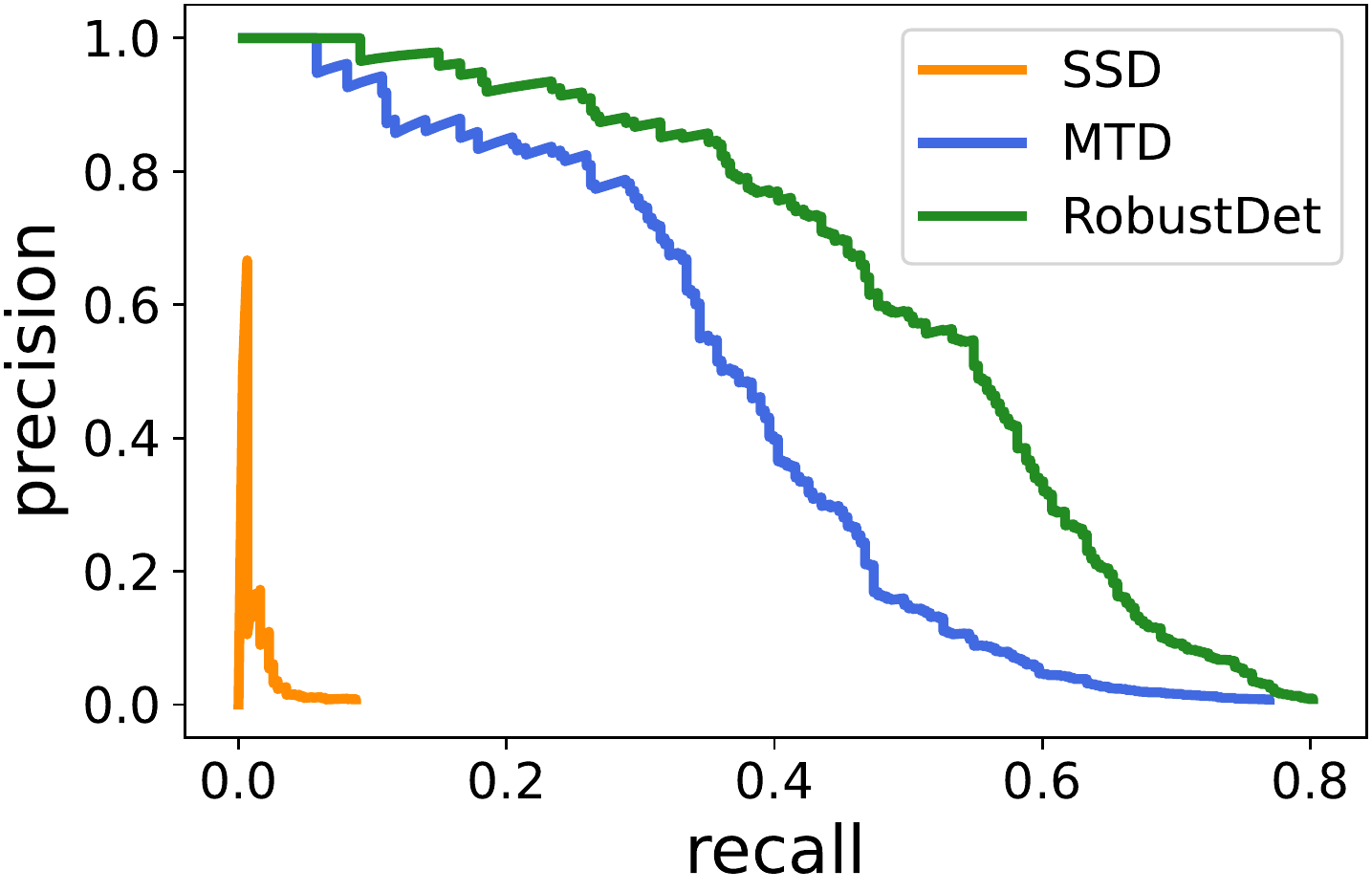}
        \end{minipage}%
        }%
        \vspace{-9pt}
        \caption{Precision-Recall (PR) curves of non-robust detector (standard SSD), and two SSD-based robust detectors, \emph{i.e.}, MTD~\cite{MTD} and our RobustDet. They are respectively evaluated under \emph{the conventional standard setting} with clean images and \emph{two detector attacks} whose adversarial images are generated from attacks of classification ($A_{cls}$) and localization ($A_{loc}$)~\cite{MTD}. It is observed that SSD has a high performance on clean images but \emph{performs rather poorly under two attacks.} The robust detector MTD is relatively robust under attacks but \emph{presents a significant performance drop on clean images}. Instead, 
        \emph{our RobustDet not only gains a reliable detection robustness on adversarial images, but also maintains a high detection performance on clean images on par with the standard SSD.} 
        }
        \label{fig:motivation}
        \vspace{-2ex}
    \end{figure}
\end{minipage}

\begin{abstract}
\vspace{-8pt}
Object detection, as a fundamental computer vision task, has achieved a remarkable progress with the emergence of deep neural networks. Nevertheless, few works explore the adversarial robustness of object detectors to resist adversarial attacks for practical applications in various real-world scenarios. Detectors have been greatly challenged by unnoticeable perturbation, with sharp performance drop on clean images and extremely poor performance on adversarial images.
In this work, we empirically explore the model training for adversarial robustness in object detection, which greatly attributes to the conflict between learning clean images and adversarial images. To mitigate this issue, we propose a Robust Detector (RobustDet) based on adversarially-aware convolution to disentangle gradients for model learning on clean and adversarial images.
RobustDet also employs the Adversarial Image Discriminator (AID) and Consistent Features with Reconstruction (CFR) to ensure a reliable robustness.
Extensive experiments on PASCAL VOC and MS-COCO demonstrate that our model effectively disentangles gradients and significantly enhances the detection robustness with maintaining the detection ability on clean images. 
Our source code and trained models are publicly available at: \href{https://github.com/7eu7d7/RobustDet}{https://github.com/7eu7d7/RobustDet}
\vspace{-3pt}

\vspace{-5pt}
\end{abstract}

\section{Introduction}\label{sec:introduction}
\begin{figure}[!t]
    \centering
        \includegraphics[width=1\linewidth]{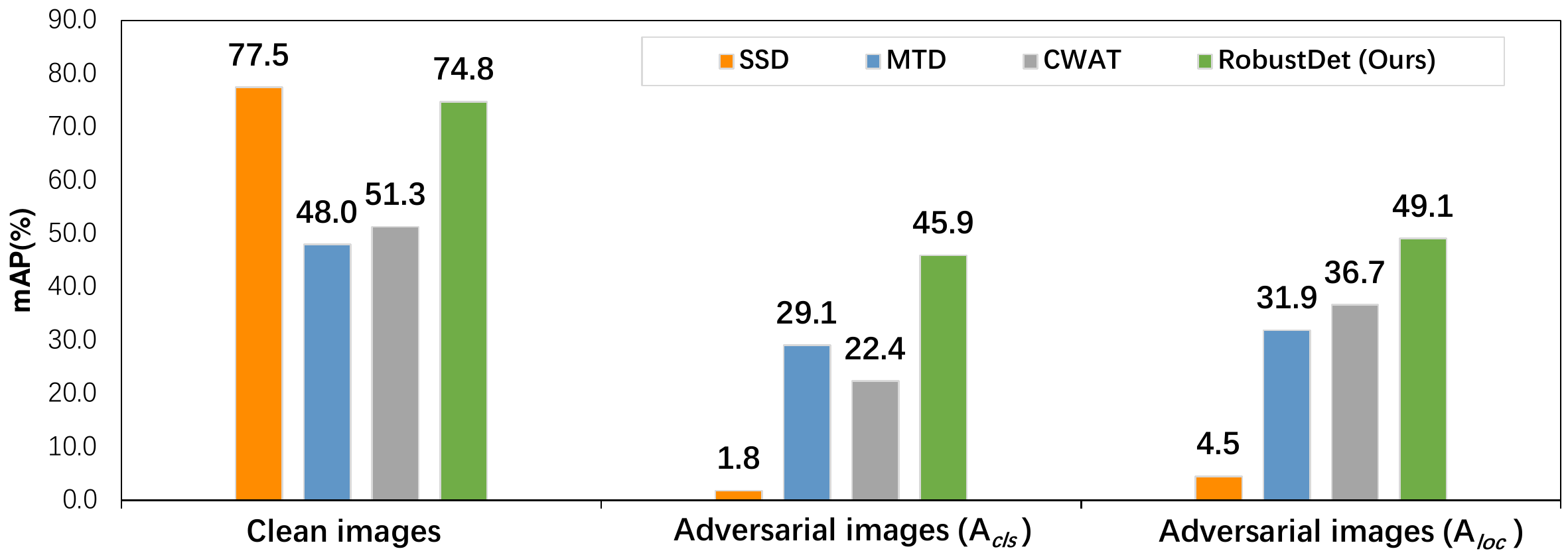}
    \caption{Detection performance comparison on clean and adversarial images for standard SSD, MTD~\cite{MTD}, CWAT~\cite{CWAT} and our RobustDet.}
    \label{fig:Motivation1-detection}
\end{figure}

Although deep neural networks (DNNs) have achieved a remarkable progress in many visual tasks such as image classification~\cite{resnet}, object detection~\cite{yolox,faster-rcnn} and semantic segmentation~\cite{pspnet,deeplabv3}, they are vulnerable to even slight, imperceptible adversarial perturbations and yield erroneous predictions~\cite{FGSM,PGD,advGAN,c-and-w}. \emph{A miss is as good as a mile.} Such vulnerability inspires increasing attentions on the adversarial robustness mainly in the image classification task~\cite{advtrain,c-and-w,High-Level,FAT,Linearization}. Nevertheless, with elaborate architectures to recognize simultaneously where and which category objects are in images,
object detectors also suffers from the vulnerable robustness and are easily fooled by adversarial attacks~\cite{DAG,UEA,ShapeShifter,CWAT,PRFA}. 
As demonstrated in \cref{fig:Motivation1-detection}, standard SSD achieves only \textbf{1.8\% mAP} on adversarial images, by \textbf{75.7\% mAP drops}! The vulnerability of object detection models seriously raises security concerns on their practicability in security-sensitive applications, \emph{e.g.}, autonomous driving and video surveillance.

\begin{figure}[!t]
    \centering
    \begin{minipage}{1\linewidth}
        \centering
        \includegraphics[width=0.42\linewidth]{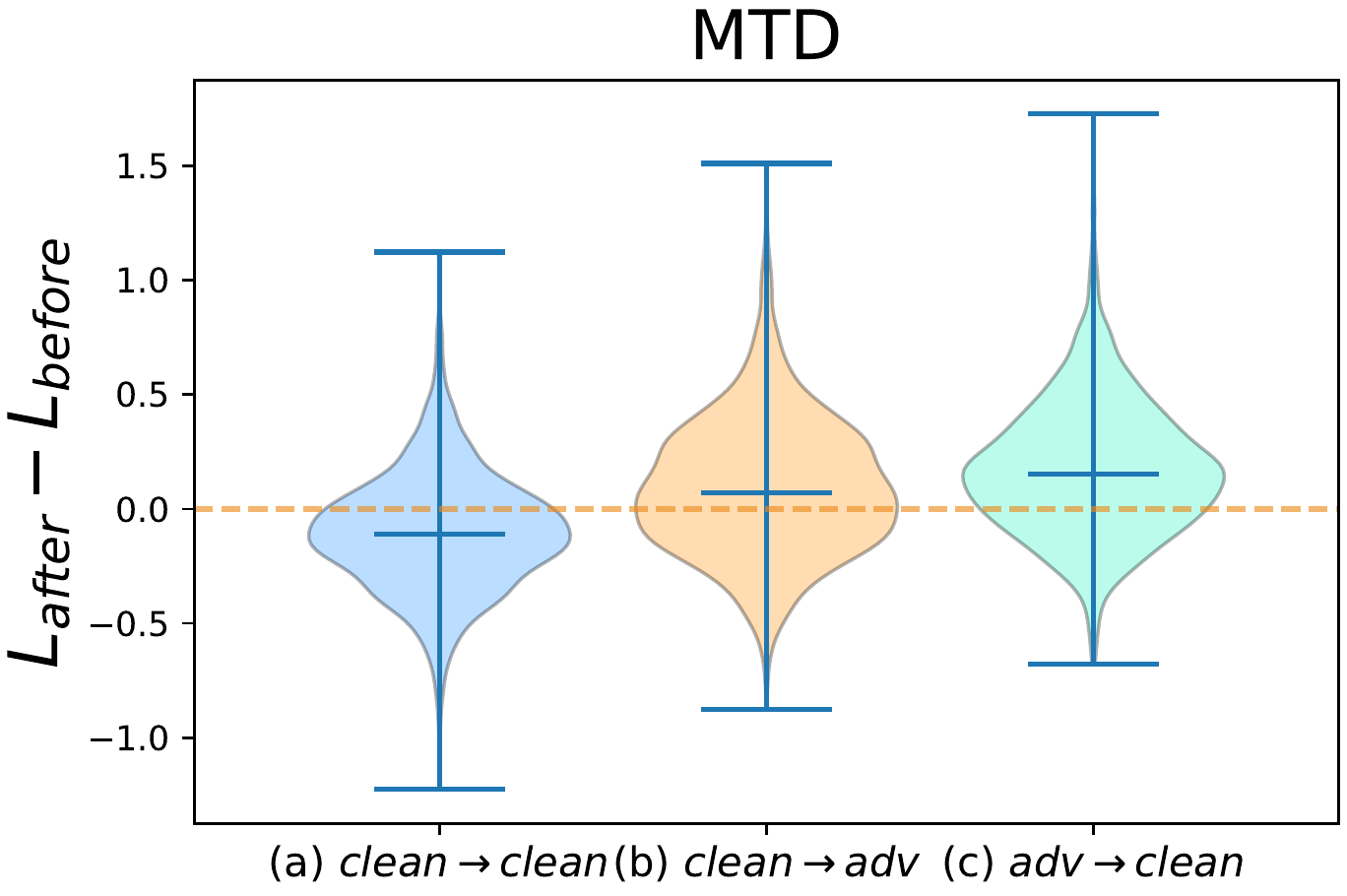}
        \hspace{10mm}
        \includegraphics[width=0.42\linewidth]{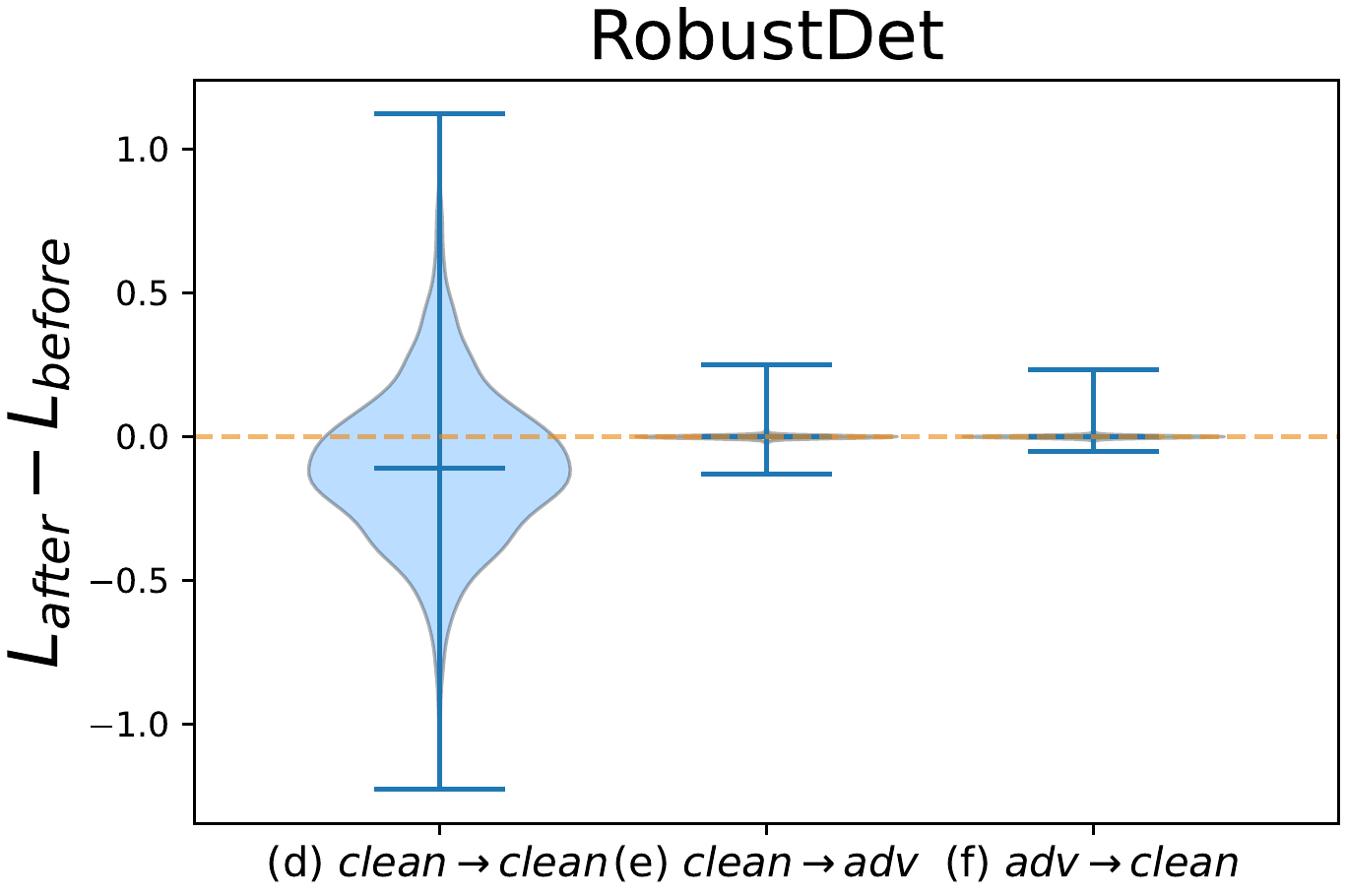}
    \end{minipage}
    \caption{
    Empirical analyses on the conflict between the learning of clean images and adversarial images via the statistics of loss changes\protect\footnotemark. 
    (a), (b) and (c) are the loss changes on robust detector MTD~\cite{MTD}. (d), (e) and (f) are the loss changes on our RobustDet.
    For both methods under $clean\rightarrow clean$, they have the decreasing loss changes for most images, indicating the favorable training. Under $clean (adv) \rightarrow adv (clean)$, MTD has the increasing loss changes for most images, indicating the inverse training effects between learning clean and adversarial images. Instead, our RobustDet has almost no effects between them, indicating a better disentanglement for learning clean and adversarial images.
    }
    \label{Figexp_conflict}
    \vspace{-6pt}
\end{figure}

The vulnerable robustness of object detectors has been impressively verified to attack two tasks of classification and localization~\cite{DAG,UEA,upset,ShapeShifter}, few researches focus on investigating the challenging countermeasure: \emph{how to defend those attacks to resist the adversarial perturbations for detectors}.
To address this issue, MTD~\cite{MTD}, as an earlier attempt, regards the adversarial training of object detection as a multi-task learning and choose those adversarial images that have the largest impact on the total loss for learning. 
Subsequently, the second related work, CWAT~\cite{CWAT}, points out the problem of class imbalance in the attack and proposes to attack each category as evenly as possible to generate more reasonable adversarial images. 
In general, these existing methods suffer from the \textbf{{detection robustness bottleneck}}: \emph{a significant degradation on clean images with only a limited adversarial robustness}, shown in~\cref{fig:motivation} and \ref{fig:Motivation1-detection}. 
That is, due to the introduction of adversarial perturbation during training, they reach a compromise for both the model accuracy on clean images and the robustness on adversarial images. This would inevitably make a concession of robust models with the performance sacrifice on clean images as well as a limited adversarial robustness for object detection.

\footnotetext{More details can be referred to our supplementary material.}
In this paper, we firstly explore the aforementioned \textbf{detection robustness bottleneck} on both clean images and adversarial images for object detection. 
Particularly, one noteworthy difference from the adversarial robustness in the image classification task, where robust models usually only have a small amount of the performance decline on clean images~\cite{Trades,IRGD}, is that robust object detectors only yields a limited robustness from adversarial training and 
suffer from a significant performance degradation by nearly 30\% on clean images (77.5\% mAP for standard SSD \emph{vs.} 48.0\% mAP for MTD~\cite{MTD} on the PASCAL VOC dataset, as shown in~\cref{fig:Motivation1-detection}). 
It indicates that, in the training phase, robust detectors hardly reach a win-win balance to trade off the robustness of adversarial images and the accuracy of clean images. 
\emph{To further investigate this issue, on one hand, we inspect the individual loss changes for both images in an adversarial robust detector}. 
A conflict between two tasks of learning clean images and adversarial images in adversarial training is observed, which can be speculated as a pitfall to explain the aforementioned detection robustness bottleneck to a certain extent. 
\emph{On the other hand, we analyze the interference between the gradients of clean images and the adversarial images for existing models}. Accordingly, strong interference is observed, indicating that an object detector has a large difficulty to distinguish no-robust and robust features. Thus, it is reasonable that models are confronted with the detection robustness bottleneck.

To mitigate this problem, we propose a Robust Detection model (RobustDet) via adversarially-aware convolution. 
The model learns different groups of convolution kernels and adaptively assigns weights to them based on the Adversarial Image Discriminator (AID).
RobustDet also employs the Consistent Features with Reconstruction (CFR) to ensure reliable robustness. By applying reconstruction constraints to make the features extracted by the model can be reconstructed as clean images as possible, the model is drived to extract more robust features for both clean and adversarial images.
Extensive experimental results on PASCAL VOC~\cite{VOC} and MS-COCO~\cite{COCO} datasets have demonstrated superior accuracy performance on clean images and promising detection robustness on adversarial images.

Overall, our contributions are summarized as follows:
\begin{enumerate}
  \item Empirically, we analyse the detection robustness bottleneck and verify the conflict between learning clean images and adversarial images for robust object detectors.
  
  \item Technically, we propose a robust detection model (RobustDet) based on adversarially-aware convolution to learn robust features for clean images and adversarial images.
  In addition, we propose Consistent Features with Reconstruction (CFR) to constrain the model to extract more robust features that can be reconstructed as clean images as possible.
  \item Experimentally, we conduct comprehensive experiments to evaluate the proposed approach for adversarial detection robustness on PASCAL VOC and MS-COCO datasets, achieving state-of-the-art performance on both clean images and adversarial images. It presents a superior accuracy performance on clean images and a promising detection robustness on adversarial images.
\end{enumerate}

\section{Related Work}

\subsection{Adversarial Attack and Defense}
\textls[-10]{For deep neural networks, their excellent feature representation capability has been demonstrated in various scenarios~\cite{resnet,VGG,MobileNetV3}. Even so, it has been criticized that neural network models easily produce totally wrong predictions under slight perturbations to inputs~\cite{advatk}. Especially, they are rather vulnerable to adversarial attacks. Accordingly, more and more adversarial attack methods have been proposed: gradient-based white box adversarial attack methods (\emph{e.g.}, FGSM~\cite{FGSM} and PGD~\cite{PGD}), and black box adversarial attack methods (\emph{e.g.}, UPSET~\cite{upset} and LeBA~\cite{LeBA}). These methods can easily fool the classification model and even a change in just one pixel would totally fool the model~\cite{one-pixel}.
To address this problem, some defense methods have been proposed~\cite{advtrain,c-and-w,High-Level,FAT,Linearization}. Among them, adversarial training is one of the most widely used and effective methods. It allows the model to continuously learn the adversarial images and focuses more on the robust features of adversarial images and clean images to ignore non-robust features.
}

\subsection{Attack and Robust Object Detector}
In recent years, seminal object detection models have been proposed, \emph{e.g.}, Faster RCNN~\cite{faster-rcnn}, SSD~\cite{SSD}, YOLOX~\cite{yolox}, and DETR~\cite{DETR}, building a series of profound and insightful milestones for object detection.
Even so, they inevitably inherit the vulnerability to attack, with the root in deep neural networks. 
Existing researches have shown that attack methods for classification tasks can also be effective in attacking object detection models~\cite{MTD}.
Object detectors have some different structures from classification models, and object detectors can be attacked more effectively for these structures. For example, DAG~\cite{DAG} and UEA~\cite{UEA} are the attack methods for object-level features by superimposing perturbation on the whole image. Dpatch~\cite{DPatch} fools the detector by adding a patch to the image. ShapeShifter~\cite{ShapeShifter} attacks detectors in the physical world. 

Instead, although attack methods for object detectors are becoming more and more efficient, there are few defense strategies in the object detection task. \cite{MTD} proposes the MTD method based on adversarial training. At each step of adversarial training, the images that can increase the loss the most are selected from the adversarial images to learn to improve the robustness of the model. \cite{CWAT} explores the problem of class imbalance in the attacks for object detectors and proposes to make the attack intensity as consistent as possible for each class. Adversarial training is performed through these images to improve the robustness of the model. 
These methods mainly focus on the generation of adversarial images and ignore the lack of robustness caused by the structure of the model. Thus, they suffer from the detection robustness bottleneck as mentioned in Sec.~\ref{sec:introduction}.

Since few research works on adversarially-robust object detectors, it is almost blind to essentially explore object detection. In this paper, we will firstly explore empirically the detection robustness bottleneck to further understand the adversarial robustness in object detection in Sec.~\ref{sec:AR}. Then, we will elaborate the proposed RobustDet to address the detection robustness bottleneck in Sec.~\ref{sec:method}. We will conduct extensive experiments to demonstrate the effectiveness of the proposed method In Sec.~\ref{sec:exp} and conclude the paper in Sec.~\ref{sec:conclusion}. 

Defenses against unseen attacks are customarily explored in classification tasks. However, for detection tasks we suffer from a lack of the most fundamental conception of their robustness. Thus, we focus more on more fundamental problems of the robustness of object detectors. Those more advanced problems need to be explored further based on this work.

\section{Adversarial Robustness in Object Detection}\label{sec:AR}

\subsection{Problem Setting}
For a clean image $x$, an object detector $f$ parameterized by $\bm \theta$, yields object bounding boxes $\{\hat{{\bm b}_i}=[p^x_i, p^y_i,w_i,h_i]\}$ with their predicted class probabilities $\{\hat{{\bm c}_i} = [\hat{c}^{bg}_i, \hat{c}^1_i, \cdots, \hat{c}^C_i] \}$ over the background ($bg$) and $C$ object categories, \emph{i.e.}, $f(x;\bm \theta) \rightarrow \{\hat{{\bm b}_i}, \hat{{\bm c}_i}\}$, where $p^x_i$ and $p^y_i$ are the coordinates of the top left corner of $\hat{{\bm b}_i}$, $w_i$ and $h_i$ are the width and height of $\hat{{\bm b}_i}$.
The localization loss $\mathcal L_{loc}=\sum_{i \in pos } L^{smooth}_1( \hat{\bm b_i}, \bm b_i )$ and the classification loss $\mathcal L_{cls}=-\sum_{i \in pos } c_{i} \log \left(\hat{c}^{u}_{i}\right)-\sum_{i \in neg} c_i \log (\hat{c}^{bg}_{i})$, 
where $\bm b_i$ is the Ground-Truth (GT) bounding box that matches the predicted bounding box $\hat{\bm b_i}$ and $\bm c_i$ denotes its GT category, and
the detection loss is $\mathcal L_{det}=\mathcal L_{loc}+\mathcal L_{cls}$.
Following MTD~\cite{MTD}, two types of attacks ($A_{cls}$ and $A_{loc}$) for object detection are specifically steered for classification and localization, respectively:
\begin{equation}
\begin{aligned}
    A_{cls}(x) &= \mathop {\arg \max }\limits_{\bar x \in {{\mathcal S}_x}} {\mathcal L}_{cls}(f(\bar x; \bm \theta),\{ {\bm c_i, \bm b_i}\} ), \\
    A_{loc}(x) &= \mathop {\arg \max }\limits_{\bar x \in {{\mathcal S}_x}} {\mathcal L}_{loc}(f(\bar x; \bm \theta),\{ {\bm c_i, \bm b_i}\} ),
\end{aligned}
\end{equation}
where $\bar x$ is the adversarial counterpart of $x$, and $\mathcal{S}_{{x}}=\left\{\bar{{x}} \cap [0,255]^{cwh} \big| \|\bar{{x}}-{x}\|_{\infty} \leq \epsilon \right\}$ is the adversarial image space centered on clean images ${x}$ with perturbation budget of $\epsilon$.
$A_{cls}$ denotes searching for the image $x$ in its $\epsilon$ neighborhood that maximizes $L_{cls}$ as the adversarial image.

\subsection{Analyses of the Detection Robustness Bottleneck}
\label{sec:challenge}
\subsubsection{(1) Conflict between Learning Adversarial images and Clean images.}
\label{sec:compromise}

To defense the attacks, 
robust models are expected to be immune to adversarial perturbations via learning shared features between clean images and adversarial images to improve the robustness of the model. This is the conventional wisdom in prevalent adversarial training for defense, especially in the image classification task. 
Nevertheless, the adversarial robustness for object detection is worrisome. Namely, robust detection models perform poorly on both clean and adversarial images, as demonstrated in Fig.~\ref{fig:motivation} and ~\cref{fig:Motivation1-detection}. In particular, 
adversarial training on both clean and adversarial images results in a significant performance drop on clean images. This may indicate a conflict between the tasks of learning clean and adversarial images; thus the model has to compromise a trade-off between adversarial and clean images.
To further explore the reasons why the model cannot learn both images well, we conduct an investigation from two aspects.

\vspace{10pt}
\textbf{Loss changes for clean and adversarial images.}
\textls[-5]{We inspect intuitively the loss changes for clean and adversarial images. Specifically, we perform a validation via $m$-step adversarial training of an adversarially-trained robust model on a batch of clean images or adversarial images and observe the loss change on another batch of images (The selection of $m$ and algorithm details are discussed in the supplementary material). 
The loss change of the adversarial (clean) image after learning the clean (adversarial) image is defined as $clean \rightarrow adv$ ($adv \rightarrow clean$). 
From the experimental results in~\cref{Figexp_conflict}, it is observed that $clean \rightarrow adv$ and $adv \rightarrow clean$ are positive for most images compared with the most negative results of $clean \rightarrow clean$. This shows that learning clean images and adversarial images will increase the loss of each other for most images. 
The impact of adversarial images on clean images is greater than that of clean images on adversarial images.
This validation shows that learning clean images and adversarial images are conflicting tasks for the model, to some extent. Thus, during the training phase, the model has the burden to well address this learning conflict.
}

\begin{figure*}[!t]
    \centering
    
    \begin{minipage}{1.05\linewidth}
        \centering
        \hspace{-20pt}
        \includegraphics[width=1\linewidth]{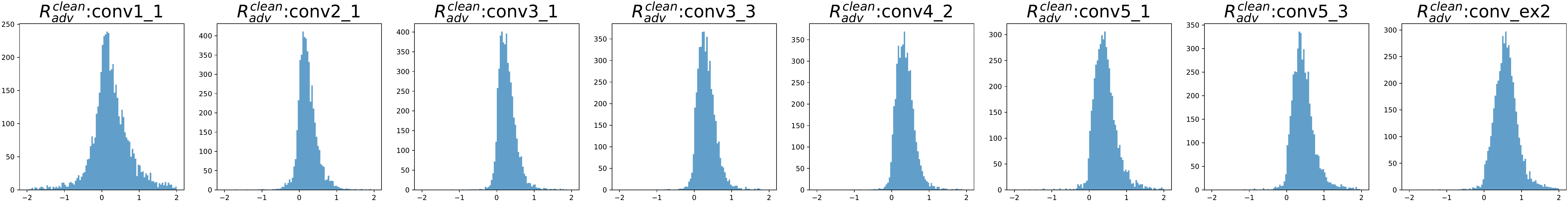}
        \\
        \hspace{-20pt}
        \includegraphics[width=1\linewidth]{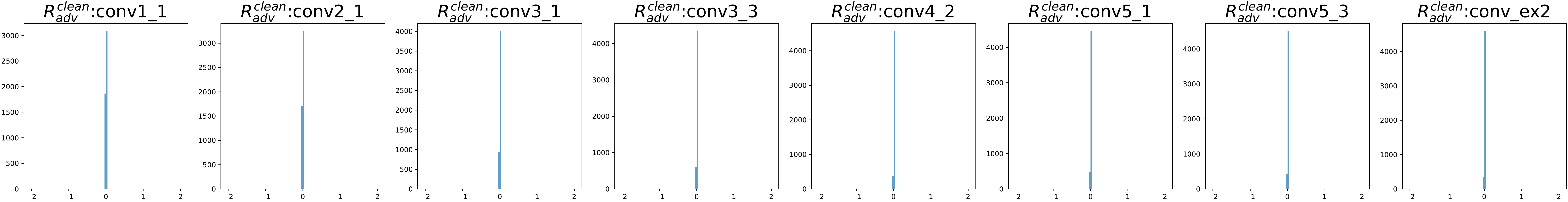}
    \end{minipage}
    
    \caption{The gradient entanglement degree $R^{clean}_{adv}$ of clean images and adversarial images based on features from different convolutional layers. The upper shows the results from SSD and the second row is from our RobustDet.}
    \label{Figexp_conflict_G}
    \vspace{-10pt}
\end{figure*}

\vspace{10pt}
\textbf{Gradient interference analysis.}
The clean image and the adversarial image are from two different domains with different patterns. There are shared features between them but also have their unique features.
A highly robust model must have parameters for extracting the shared features and another two part parameters for extracting unique features that are orthogonal to each other.
For an adversarial trained robust model, the shared features of two kinds of images should have been well learned, and only the part processing unique features still needs reinforcement.
Therefore, for this model, the gradients generated by the two kinds of images should have low correlation and be nearly orthogonal.

\textls[-5]{Accordingly, we define the intensity of gradient entanglement: $\mathcal{R}^{g_1}_{g_2}={{g_1}^{T}g_2}/{|{g_2}|^2}$,
where $g_1$ and $g_2$ are the gradient vectors of the two kind of images.
The greater the gradient entanglement of the two kinds of images, the more serious the interference between them and the model does not distinguish the unique features well.
Based on the experimental results of the above loss variations, the greater the gradient entanglement, the more difficult the conflict between the two kinds of images can be reconciled. The smaller gradient entanglement indicates that the model has enough ability to distinguish the shared features from their unique features and can disentangle the clean images and the adversarial images.
It can be seen from the~\cref{Figexp_conflict_G} that the gradient entanglement between the clean image and the adversarial image on the adversarial trained robust model is quite high, and even negative values appear in the first few layers. This shows that the updated directions of some clean images and adversarial images on the adversarial-trained model are completely opposite, which also indicates the conflict between the two kinds of images. 
When learning one kind of image, it will inevitably have an impact on another kind of image, which leads to a detection robustness bottleneck.
}

\begin{figure}[!t]
    \centering
    \begin{minipage}{1.1\linewidth}
        \centering
        \hspace{-50pt}
        \includegraphics[width=0.52\linewidth]{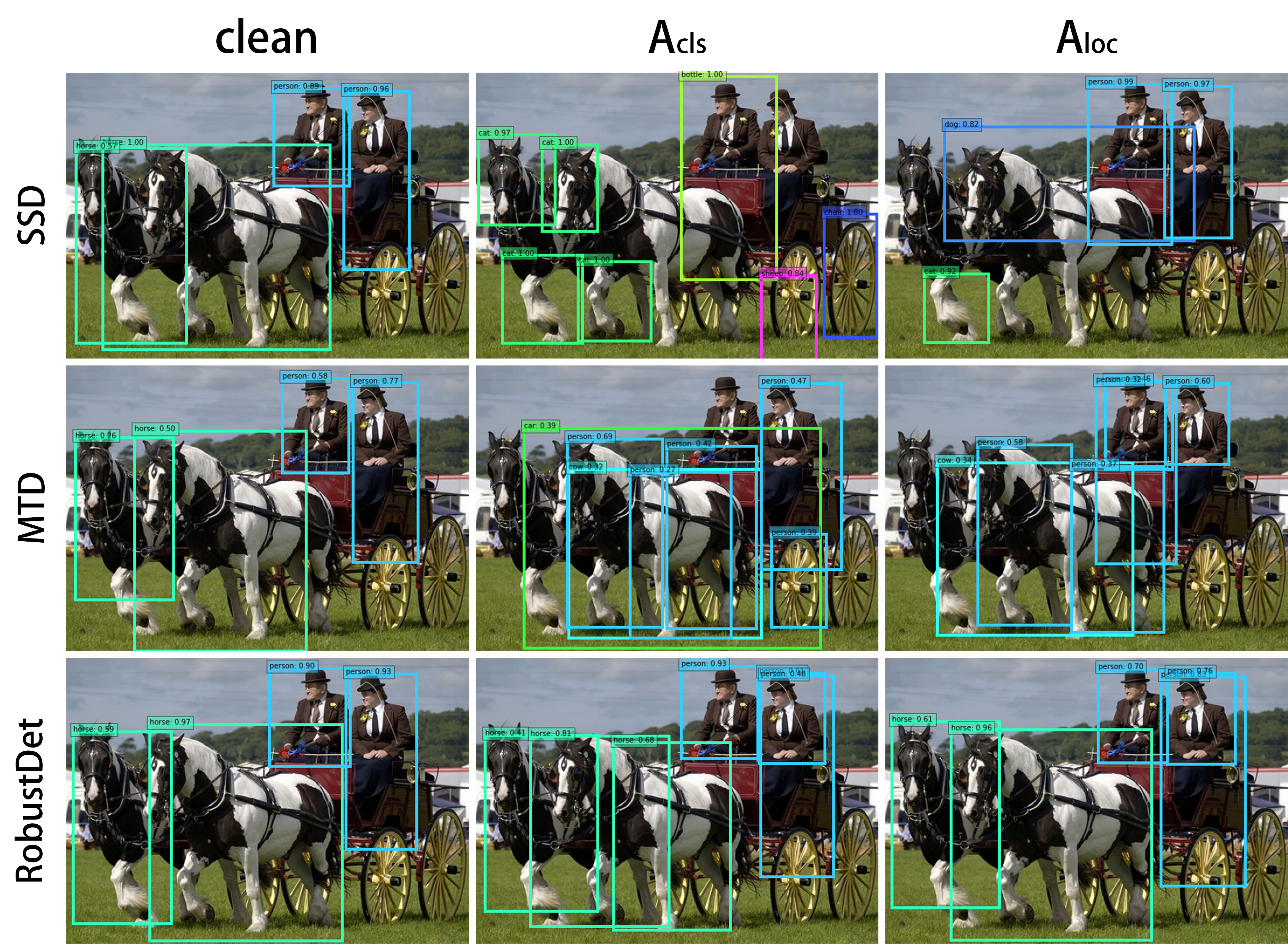}
        \quad \quad
        \hspace{-10pt}
        \includegraphics[width=0.38\linewidth]{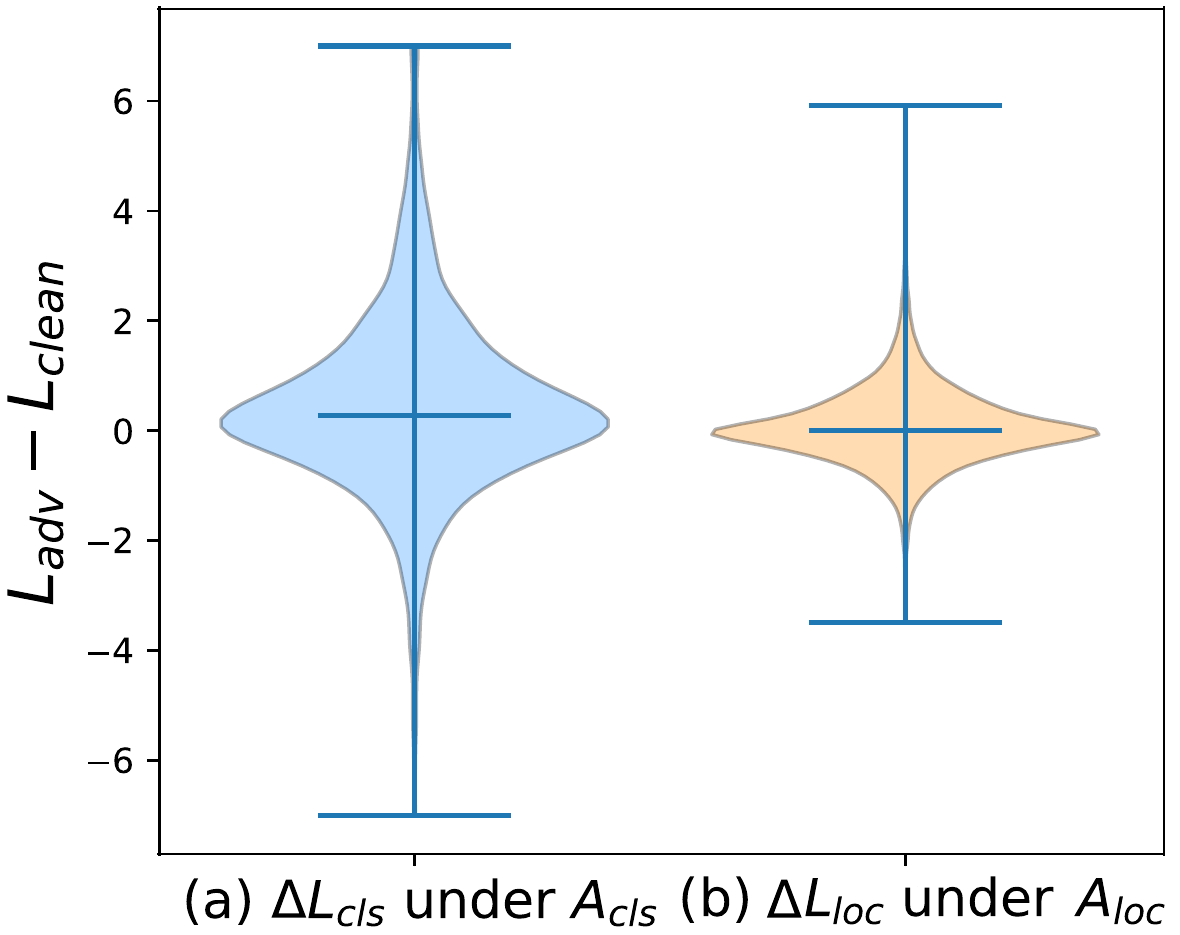}
    \end{minipage}
    \caption{\textbf{Left:} The detection results of the standard SSD, MTD and our RobustDet on the clean image and two adversarial images attacked from classification ($A_{cls}$) and localization ($A_{loc}$). MTD and RobustDet are robust models taking SSD as their base-models.
    \textbf{Right:} Under the attacks of $A_{cls}$ and $A_{loc}$, the corresponding loss changes between the adversarial image and the original image.
    }
    \vspace{-3ex}
    \label{Fig_cls_loc}
\end{figure}

\vspace{-5pt}
\subsubsection{(2) The Conflict to the Robustness of Classification and Location.}
$\\$

\vspace{-1.12em}
\noindent
We compare the detection results of the non-robust model and the adversarial trained robust model on the clean image, the $A_{cls}$ adversarial image, and the $A_{loc}$ adversarial image.
It can be seen from the~\cref{Fig_cls_loc} that the non-robust model will locate the wrong object with high confidence when applying an attack.
The robust model will not completely confuse to the attack, but its classification and localization accuracy on both clean images and adversarial images have greatly decreased.
The robustness of localization objects is much better than classification. It can be seen from the figure that the bounding boxes predicted by the robust models do not have as large deviations as the classification.

\textls[-5]{The results in~\cref{Fig_cls_loc} shows that the variation of $L_{cls}$ when applying $A_{cls}$ attack compared to the variation of $L_{loc}$ when applying $A_{loc}$ attack is much larger. This also indicates that the classification module is less robust and more vulnerable to attack.
These both shows that the conflict between the two images under the classification subtask is more serious than the localization subtask. The scores given in the classification part will also determine the selection of the bounding box. Therefore, this conflict will further damage the performance of the model.
}

\vspace{-6pt}
\section{Methodology}
\vspace{-4pt}

\label{sec:method}

\subsection{Overall Framework}

\begin{figure*}[!t]
    \centering
    \begin{minipage}{1.08\linewidth}
        \hspace{-16pt}
	    \includegraphics[width=1\linewidth]{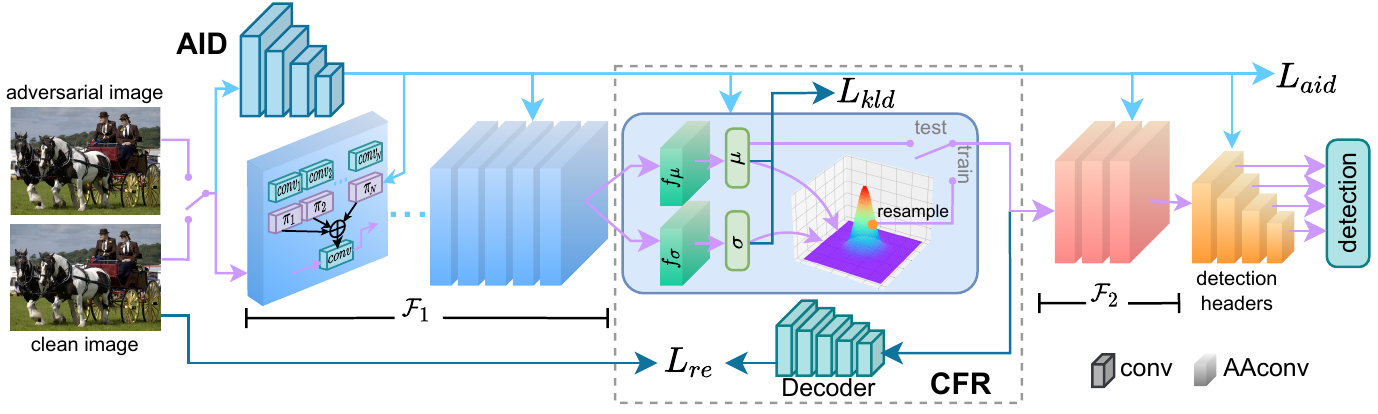}
    \end{minipage}
    \vspace{-6pt}
    \caption{The overall architecture of RobustDet based on SSD. 
    The CFR is inserted into the SSD backbone followed by the first detection layer (conv4\_3), and the two parts in front and behind this layer are named $\mathcal F_1$ and $\mathcal F_2$.
    The \textcolor[rgb]{0.4, 0.8, 1.0}{blue} arrows are the data flow of AID, whose outputs are used as weights of AAconv. The \textcolor[rgb]{0.8, 0.6, 1.0}{purple} arrows are the primary data flow when RobustDet detect objects. The \textcolor[rgb]{0.0627, 0.451, 0.62}{teal} arrows are the reconstruction data flow during training.
    }
    \vspace{-10pt}
    \label{Fig_struct}
\end{figure*}

Based on the aforementioned analyses in~\cref{sec:challenge}, the conflict between the learning of clean and adversarial images has adversary effects on the robustness of classification and localization. 
To address this problem, we propose a RobustDet model for defenses against adversarial attacks (\cref{Fig_struct}).
We detect objects through adversarially-aware convolution and use an Adversarial Image Discriminator (AID) to generate the weights for the adversarially-aware convolution kernel based on the perturbations of the input image.
Furthermore, inspired by VAE~\cite{VAE}, the image reconstruction constraint via CFR is considered for reconstructing images as clean images to facilitate the model to learn robust features.

\vspace{-3pt}
\subsection{Adversarially-Aware Convolution (AAconv)}
Existing models essentially utilize the shared model parameters for learning adversarial images and clean images.
This inevitably makes the model suffer from a detection robustness bottleneck.
There are objective distinctions between adversarial images and clean images. Admitting these distinctions rather than forcing the detector to learn these two images with the same parameters would be a better choice.
Making the model explicitly distinguish these two kinds of images and detect them with different parameters will alleviate the conflict between these tasks.
Inspired by~\cite{dconv}, we propose adversarially-aware convolution in our RobustDet model to learn robust features for clean images and adversarial images.

RobustDet employs different kernels to convolve clean images and adversarial images.
Different parameters will be used for different perturbed images.
The generation of the convolution kernel is controlled by an adversarial image discriminator $D$.
Before the model detects objects in an image, the adversarial image discriminator $D$ will first generate the $M$-dimensional probability vector of the image $\mathcal{P}=D(x)=\{\pi_1,\pi_2,...,\pi_M\}$.
This probability vector is used as the weights to control the convolution kernels generation.
Then the parameters of the finally generated convolution kernel can be write as: $\dot{\theta}^{AAconv} = \sum_{i=1}^{M} \theta^{AAconv}_{i} \cdot \pi_i$, where $\theta^{AAconv}_i$ denotes parameters of dynamic convolution kernels in our AAconv module, where $i$ indicates the index of $i$-th convolution kernel.

RobustDet uses adversarially-aware convolutions to adaptively detect different images with different kernels and thus it can effectively learn robust features for clean and adversarial images.
It not only extracts the shared features, but also can be responsible for specific features for clean and adversarial images.
Therefore, it is more effective to alleviate the detection robustness bottleneck.

\vspace{-3pt}
\subsection{Adversarial Image Discriminator (AID)}
The generation of the adversarially-aware convolution kernels is controlled by the adversarial image discriminator. And this module may also be attacked and give the wrong weight. Wrong weights will lead the wrong convolution kernel to be generated,
which will be a disaster for the model. 
Accordingly, in order to improve its robustness, we employ Online Triplet Loss~\cite{Triplet} to the adversarial image discriminator.
Specifically, we consider the probability distribution of the same kind of images (\emph{i.e.}, clean or adversarial images) as close as possible and the different kinds of images (clean or adversarial images) as far away as possible. A margin between the probability distributions of the two kinds of images outputs is introduced to strengthen the robustness of the adversarial image discriminator. 
Jensen-Shannon (JS) divergence~\cite{JSD} is utilized to measure the distance between two probability distributions, $P_1$ and $P_2$ (two distributions as an example for JS divergence): 
$JS\left(P_{1} \| P_{2}\right)=\frac{1}{2} KL\left(P_{1} \| \frac{P_{1}+P_{2}}{2}\right)+\frac{1}{2} KL\left(P_{2} \| \frac{P_{1}+P_{2}}{2}\right)$. Overall, the AID loss is defined as follows, 
\begin{equation}
\begin{aligned}
\mathcal{L}_{aid}=\textstyle\sum_{i=1}^{N_T}\left[JS\left(D\left(x_{i}^{a}\right) \| D\left(x_{i}^{p}\right)\right)- \right.  \left. JS\left(D\left(x_{i}^{a}\right) \| D\left(x_{i}^{n}\right)\right)+\gamma\right]_{+},
\end{aligned}
\end{equation}
where $x^p$ ($x^n$) is a randomly selected image from one minibatch that has the same (opposite) type (\emph{i.e.}, clean or adversarial image) as the anchor instance $x^a$ in one triplet, $\gamma$ is the margin between $x^n$ and $x^p$, $N_T$ is the number of triplets, and $[\cdot]_+$ clips values to $[0,+\infty]$.

\vspace{-3pt}
\subsection{Consistent Features with Reconstruction (CFR)}
To alleviate the negative effects of adversarial perturbation, our RobustDet aims to ensure the feature distribution of an adversarial image in the neighbourhood of its clean image.
Thus, inspired by VAE~\cite{VAE}, our RobustDet reconstructs consistent features of clean/adversarial images with clean images via our AAconvs. Assume that the output feature map of the convolutional layer after the conv4\_3 layer (VGG backbone) comes from a multivariate Gaussian distribution with an diagonal covariance matrix $\mathcal{N}(\bm \mu = (\mu_1,...,\mu_N), \bm \Sigma = diag(\sigma^2_1,...,\sigma^2_N))$. For simplicity, $\bm \sigma=(\sigma^2_1,...,\sigma^2_N)$.
Instead of directly predicting the features that are ultimately used for detection, our model predicts the mean $\bm \mu$ and standard deviations $\bm \sigma$ of its feature distribution: $\bm \mu=f_{\bm \mu}(\mathcal F_1(x))$, $\bm \sigma=f_{\bm \sigma}(\mathcal F_1(x))$, where $f_{\bm \mu}$ and $f_{\bm \sigma}$ are the two layers of the model that predict the mean and standard deviations, $\mathcal F_1(x)$ and $\mathcal F_2(x)$ is two parts of VGG that split by conv4\_3.
From this distribution, a $N$-dimensional feature vector is randomly sampled as the robust feature for the input image, which is used for subsequent CFR and object detection in the training phase.
Then the reconstruction loss can be defined as:
\begin{equation}
\begin{aligned}
\mathcal{L}_{re}&=\|G\left(\bm z\right)-x\|^2, \quad \bm z \sim \mathcal{N}(\bm \mu, \bm \Sigma), \\
\end{aligned}
\end{equation}
where $\| \cdot \|^2$ indicates $\ell_{2}$ norm, and $x$ is the clean image. Once this feature distribution is learnt, our model can generate the similar features for an adversarial image and its clean counterpart image. Thus, 
in the testing phase, the predicted mean $\mu$ is directly used as the robust feature for detection. 

Furthermore, similar to VAE, we also have an additional constraint to prevent the predicted distribution from collapse (\emph{e.g.}, $\bm \mu$ and $\bm \sigma$ are approximate to zero):
\begin{equation}
\begin{aligned}
\mathcal{L}_{kld}&=\textstyle\sum^{N}_{i=1} \frac{1}{2N}\left(-\log \sigma_i^{2}+\mu_i^{2}+\sigma_i^{2}-1\right),\\
\end{aligned}
\end{equation}

Overall, the total loss of our RobustDet is summarized as follows,
\begin{equation}
\mathcal{L}=\beta (\mathcal{L}_{det}+a\mathcal{L}_{aid})+b\mathcal{L}_{re}+c\mathcal{L}_{kld},
\end{equation}
where $\beta$, $a$, $b$ and $c$ are the hyper-parameters.

\vspace{-4pt}
\section{Experiments}
\vspace{-4pt}

\label{sec:exp}

\subsection{Implementation Details}
Our experiments are conducted on PASCAL VOC~\cite{VOC} and MS-COCO~\cite{COCO} datasets. Mean average precision (mAP) with IoU threshold 0.5 is used for evaluating the performance of standard and robust models.

\textls[0]{The proposed method is rooted in the one-stage detector SSD~\cite{SSD} with VGG16 as the backbone. Considering that Batch Normalization would increase the adversarial vulnerability~\cite{benz2021batch}, we make a modification on VGG16 without batch normalization layers~\cite{SSD}. \textcolor{black}{In experiments, we use the model pre-trained on clean images for adversarial training and employ Stochastic Gradient Descent (SGD) with a learning rate of $10^{-3}$, momentum 0.9, weight decay 0.0005 and batch size 32 with the multi-box loss.} 

For the robustness evaluation, we follow the same setting to MTD~\cite{MTD} and CWAT~\cite{CWAT} for a fair comparison and use three different attacks, PGD~\cite{PGD}, CWA~\cite{CWAT} and DAG~\cite{DAG}. Among them, CWA and DAG are specifically designed for object detectors.
For adversarial training, we also follow the same attack setting to MTD~\cite{MTD} and CWAT~\cite{CWAT} for a fair comparison; namely, we use the PGD-20 attacker with budget $\epsilon=8$ to generate adversarial examples~\cite{MTD}. 
And we set the margin in $\mathcal L_{aid}$ as $\gamma=0.6$ and $N_T$ is calculated from the mini-batch, and hyper-parameters in $\mathcal L$ as $\beta=0.75$, $a=3$, $b=0.16$ and $c=5$. RobustDet* represents RobustDet with CFR.
}

\subsection{Detection Robustness Evaluation}
In this section, we evaluate the proposed method in comparison with the state-of-the-art approaches on the PASCAL VOC and MS-COCO datasets in \cref{tab:VOC} and \ref{tab:COCO}. The scenarios in MS-COCO are more complex than PASCAL VOC, and thus it is also more challenging to make the model robust on this dataset.
Considering that the object detector has two tasks of classification and localization, we can use PGD to attack the classification ($A_{cls}$) and localization ($A_{loc}$). For DAG attacks, we perform 150 steps to make an effective attack. The experimental results are provided in \cref{tab:VOC} and \cref{tab:COCO}.

\begin{table}[t!]
\centering
\setlength\tabcolsep{7.3pt} 
\renewcommand{\arraystretch}{1.12}
\caption{The evaluation results using various adversarial attack method on PASCAL VOC 2007 test set\protect\footnotemark. 
}
\begin{tabular}{lllllll}
\toprule
\textbf{Method} & \textbf{Clean} & $\bm{A_{cls}}$  & $\bm{A_{loc}}$ & \textbf{CWA} & \textbf{DAG} \\
\midrule
SSD & \textbf{77.5} & 1.8 & 4.5 & 1.2 & 4.9 \\
SSD-AT($A_{cls}$)~\cite{MTD} & \UT{46.7}{\Pless{30.8}} & \UT{21.8}{\Pmore{20.0}} & \UT{32.2}{\Pmore{27.7}} & - & \UT{28.0}{\Pmore{23.1}} \\
SSD-AT($A_{loc}$)~\cite{MTD} & \UT{51.9}{\Pless{25.6}} & \UT{23.7}{\Pmore{21.9}} & \UT{26.5}{\Pmore{22.0}} & - & \UT{17.2}{\Pmore{12.3}} \\
MTD~\cite{MTD} & \UT{48.0}{\Pless{29.5}} & \UT{29.1}{\Pmore{27.3}} & \UT{31.9}{\Pmore{27.4}} & \UT{18.2}{\Pmore{17.0}} & \UT{28.5}{\Pmore{23.6}} \\
CWAT(PGD-10)~\cite{CWAT} & \UT{51.3}{\Pless{26.2}} & \UT{22.4}{\Pmore{20.6}} & \UT{36.7}{\Pmore{32.2}} & \UT{19.9}{\Pmore{18.7}} & \UT{50.3}{\Pmore{45.4}} \\ \hline
\textbf{RobustDet} (\textbf{ours}) & \UT{75.4}{\Pless{2.1}} & \UT{41.5}{\Pmore{40.0}} & \UT{45.2}{\Pmore{40.7}} & \UT{42.4}{\Pmore{41.2}} & \UT{52.0}{\Pmore{47.1}} \\
\textbf{RobustDet*} (\textbf{ours}) & \UT{74.8}{\Pless{2.7}} & \UT{\textbf{45.9}}{\Pmore{44.1}} & \UT{\textbf{49.1}}{\Pmore{44.6}} & \UT{\textbf{48.0}}{\Pmore{46.8}} & \UT{\textbf{56.6}}{\Pmore{51.8}} \\
\bottomrule
\end{tabular}
\vspace{-12pt}
\label{tab:VOC}
\end{table}
\footnotetext{\Pless{} and \Pmore{} indicate the mAP decrease or increase compared with the baseline SSD, respectively. '-' indicates the result is not provided in the existing work.}

\begin{table}[t!]
\centering
\setlength\tabcolsep{7.3pt}
\caption{The evaluation results using various adversarial attack method on MS-COCO 2017 test set.}
\begin{tabular}{llllll}
\toprule
\textbf{Method} & \textbf{Clean} & $\bm{A_{cls}}$  & $\bm{A_{loc}}$ & \textbf{CWA} & \textbf{DAG}   \\
\midrule
SSD                      & \textbf{42.0} & 0.4  & 1.8  & 0.1  & 8.1   \\
MTD~\cite{MTD}                        & \UT{24.2}{\Pless{17.8}}  & \UT{13.0}{\Pmore{12.6}}  & \UT{13.4}{\Pmore{11.6}}  & \UT{7.7}{\Pmore{7.6}}   & -     \\
CWAT(PGD-10)~\cite{CWAT}               & \UT{23.7}{\Pless{18.3}}  & \UT{14.2}{\Pmore{13.8}}  & \UT{15.5}{\Pmore{13.7}}  & \UT{9.2}{\Pmore{9.1}}   & -     \\
\midrule
\textbf{RobustDet} (\textbf{ours})                  & \UT{36.7}{\Pless{5.3}} & \UT{\textbf{20.6}}{\Pmore{20.2}} & \UT{\textbf{19.4}}{\Pmore{17.6}} & \UT{\textbf{20.5}}{\Pmore{20.4}} & \UT{\textbf{24.5}}{\Pmore{16.4}}  \\
\textbf{RobustDet*} (\textbf{ours})             & \UT{36.0}{\Pless{6.0}} & \UT{20.0}{\Pmore{19.6}} & \UT{19.0}{\Pmore{17.2}} & \UT{19.9}{\Pmore{19.8}} & \UT{16.5}{\Pmore{8.4}} \\
\bottomrule
\end{tabular}
\vspace{-10pt}
\label{tab:COCO}
\end{table}

\textls[-5]{
In \cref{tab:VOC} and \cref{tab:COCO}, under different datasets, in compare with standard SSD, MTD (rooted in SSD) suffers from a significant performance degradation on clean images while gaining limited robustness.
For example, on the PASCAL VOC dataset, its mAP performance on clean images significantly drops from 77.5\% to 1.8\% and 4.5\% under $A_{cls}$ and $A_{loc}$ attacks, respectively. 
It also exhibits a poor robustness under CWA and DAG attacks with only 1.2\% and 4.9\% mAP, respectively. Besides, as for existing robust methods, MTD and CWAT only gain less than 30\% mAP under $A_{cls}$ and 40\% under $A_{loc}$ and even lose almost 30\% mAP on clean images compared with baseline SSD. 
Instead, our proposed RobustDet not only obtains a high robustness on adversarial images, but also ensures a comparable performance with standard SSD on clean images with a slight performance decrease. On the PASCAL VOC dataset, RobustDet obtains larger than 40\% mAPs on adversarial images to defense detection attacks and just loses 2.7\% at most on clean images, in comparison with standard SSD. Besides, it also presents a remarkable performance on the MS-COCO dataset. For instance, 
RobustDet achieves 24.5\% under the DAG attack with only 6\% mAP decline at most on clean images (RobustDet 36.7\% vs. SSD 42.0\%).
}

\vspace{-5pt}
\subsection{Model Evaluation and Analysis}

\begin{table}[t!]
\centering
\setlength\tabcolsep{11.4pt}
\caption{The ablation study of our model under various adversarial attack method on PASCAL VOC 2007 test set.}
\begin{tabular}{llllll}
\toprule
\textbf{Method} & \textbf{Clean} & $\bm{A_{cls}}$  & $\bm{A_{loc}}$ & \textbf{CWA} & \textbf{DAG}   \\
\midrule
RobustDet w/o $L_{aid}$          & 74.9 & 37.3 & 44.9 & 37.9 & 51.8  \\
RobustDet* w/o $L_{re}$ & 74.6 & 27.5 & 41.8 & 28.6 & 55.9 \\
RobustDet                   & \textbf{75.4} & 41.5 & 45.2 & 42.4 & 52.0 \\
RobustDet*               & 74.8  & \textbf{45.9} & \textbf{49.1} & \textbf{48.0} & \textbf{56.6} \\
\bottomrule
\end{tabular}
\vspace{-10pt}
\label{tab:Ablation_study}
\end{table}

\noindent
\textbf{Ablation Study on $L_{aid}$.}
The adversarial image discriminator may also be attacked. Thus, the AID loss is introduced to improve its robustness. As shown in \cref{tab:Ablation_study},
without $L_{aid}$ RobustDet has a performance decrease on both clean and adversarial images, especially on adversarial images. For instance, it drops by 4.2\% mAP under the $A_{cls}$ attack and by 4.5\% mAP under the CAW attack. 
The absence of $L_{aid}$ makes it easier for AID to confuse clean and adversarial images.

\noindent
\textbf{Ablation Study on Consistent Features with Reconstruction.}
We compare RobustDet (without CFR) and RobustDet* (with CFR) for the ablation study on CFR. 
On the PASCAL VOC dataset, as shown in \cref{tab:VOC} and \ref{tab:Ablation_study}, the detection robustness has been improved with the CFR module by 4.1\% gains at least (RobustDet 47.5\% vs. RobustDet* 45.9\% under CWA attack) and by 5.6\% at most (RobustDet 42.4\% vs. RobustDet* 46.8\% under CWA attack). On MS-COCO, \cref{tab:COCO} shows that RobustDet* has a lower performance than RobustDet under all the attacks.
This reconstruction can be treated as VAE in VGG-16 whose capacity is relatively limited to learn so many categories, thus compromising the overall training of the model and leading to the performance degradation.  
Besides, CFR has two losses of $L_{kld}$ and $L_{re}$. In~\cref{tab:Ablation_study}, without $L_{re}$, RobustDet* has a significant decrease under attacks with similar performance on clean images, compared with the baseline. This indicates the model cannot effectively predict both samples into the same distribution.

\vspace{-15pt}
\subsubsection{Attack using Different PGD Steps.}

\begin{figure}[t]
\centering
    \begin{minipage}{1.125\textwidth}
    \hspace{-15pt}
	\subfloat[$A_{cls}$ PGD attack]{
	\includegraphics[width = 0.295\textwidth]{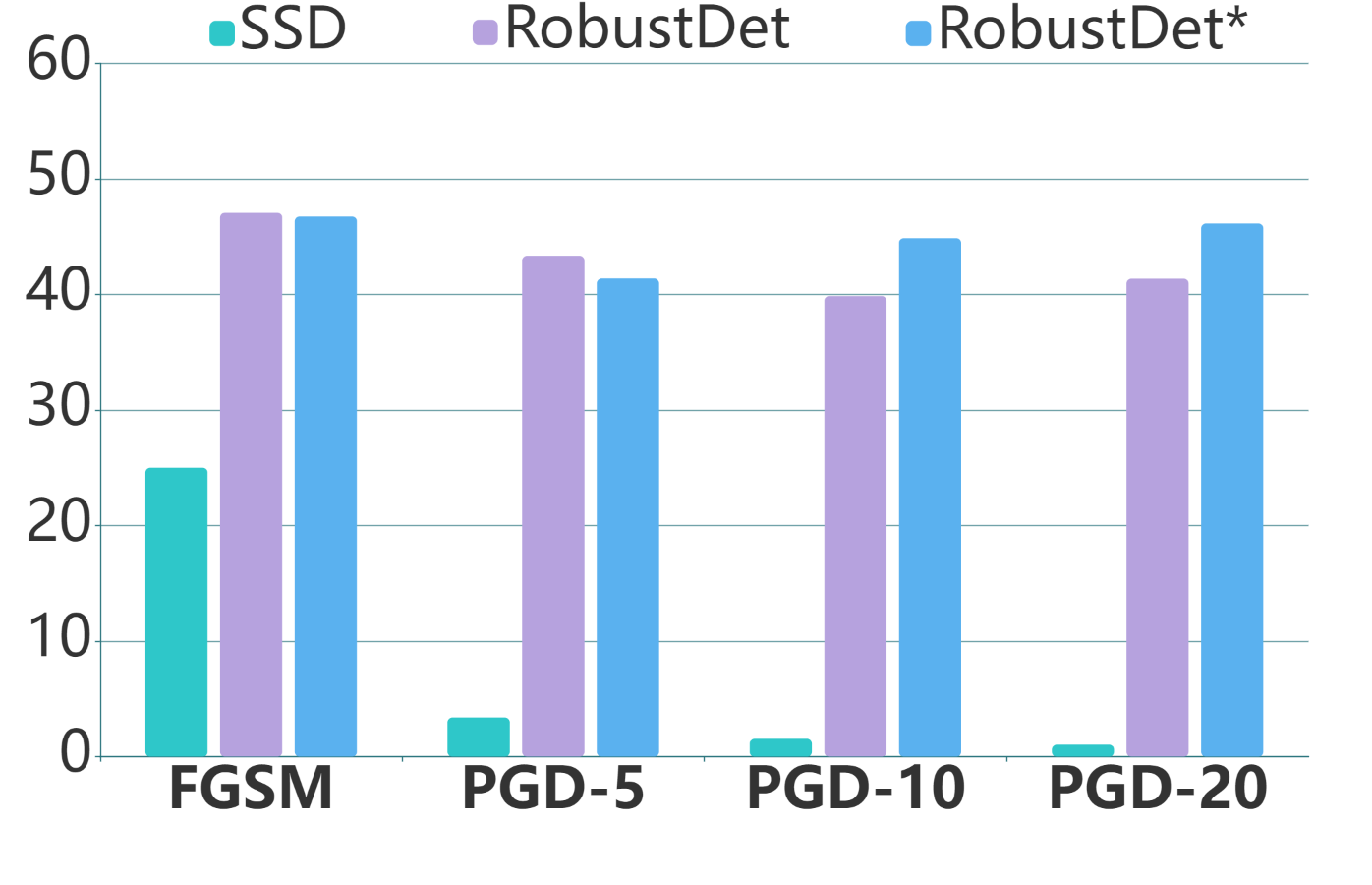}
	}
	\hspace{-10pt}
	\subfloat[$A_{loc}$ PGD attack]{
	\includegraphics[width = 0.295\textwidth]{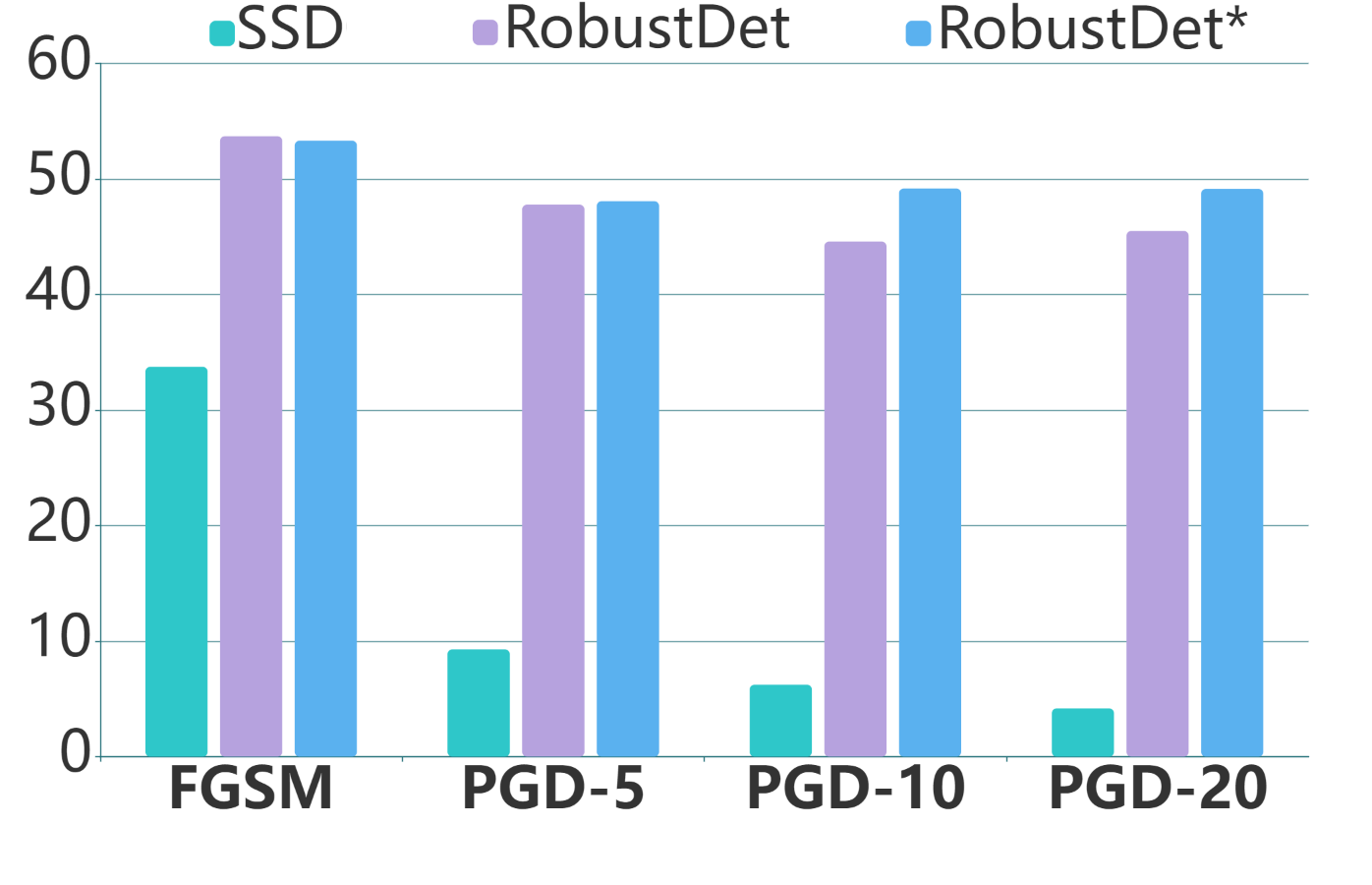}
	}
	\hspace{-15pt}
	\subfloat[Confidence distribution]{
	\includegraphics[width=0.35\linewidth]{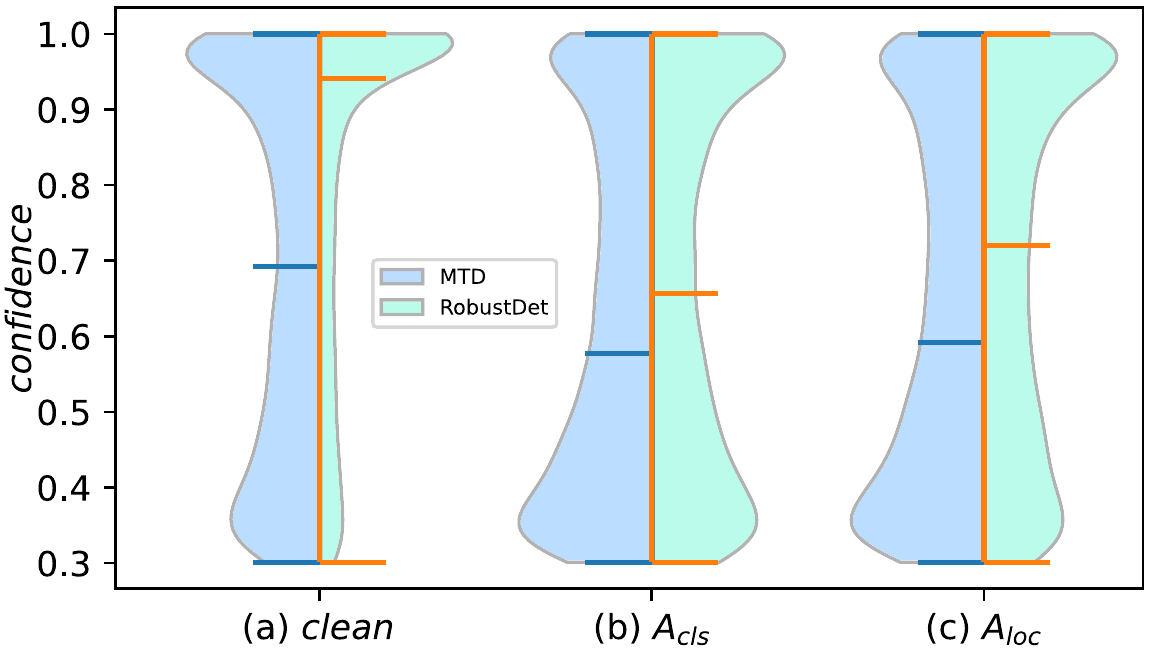}
	}
	\end{minipage}
	\vspace{-5pt}
\caption{(a) and (b): The robustness of our model under attacks with $\epsilon = 8$ using different PGD steps. (c): Under the attack on $L_{cls}$ and $L_{loc}$ loss, the corresponding loss changes between the adversarial image and the original image.}
\label{fig_steps}
\vspace{-10pt}
\end{figure}

To verify the generalization ability of our model against different steps of PGD attacks, we follow the setting of MTD~\cite{MTD} and provide the performance of the model under various steps of PGD attack on PASCAL VOC in~\cref{fig_steps}(a) and (b). For non-robust SSD, the performance decreases dramatically with the increase of iteration steps. Our model shows a strong robustness under a variety of PGD attacks with different number of steps. 
Combined with the experimental results of CWA and DAG in~\cref{tab:VOC}, it shows that our model has a promising generalization ability and can defend well even if the attacks are somewhat different from the training.

\noindent
\textbf{Analysis on Gradient Disentanglement.}
\textls[-5]{As discussed in \cref{sec:compromise}, the detection robustness bottleneck can attribute to a conflict between learning adversarial images and clean images. 
It can be observed from \cref{Figexp_conflict}, a adversarially-trained SSD model that learns adversarial (clean) images have a negative impact on the learning of clean (adversarial) images, making the loss increase. But a adversarially-trained RobustDet has almost no similar impact. The average loss variation is less than 0.1. It is also evidenced from \cref{Figexp_conflict_G} that the gradients of RobustDet on both samples are almost orthogonal. 
These indicate RobustDet can effectively alleviate the detection robustness bottleneck and learn both images better.}

\noindent
\textbf{Analysis on Confidence Distribution.}
To further verify our RobustDet addressing the conflict, 
the confidence distribution of bounding boxes that the robust model MTD and our RobustDet produce on clean and adversarial images($A_{cls}$ and $A_{loc}$), respectively, in \cref{fig_steps}(c). Here we set the filtering threshold for the confidence of the bounding box to be 0.3. From which it is evident that the confidence of the MTD robust model on both the clean and adversarial images is quite low (around 0.7 on clean, 0.6 on $A_{cls}$ and $A_{loc}$), which is also a manifestation of conflict. In contrast, the confidence of our proposed RobustDet model is fairly high on clean images (around 0.95, by 0.25 higher than MTD) and the confidence on adversarial images is mostly distributed in the higher part (around 0.65 on $A_{cls}$ and 0.7 on $A_{loc}$). This result can also well illustrate that our method can effectively alleviate the conflict and the detection robustness bottleneck.

\vspace{-8pt}
\section{Conclusion}
\vspace{-6pt}

\label{sec:conclusion}

In this work, we investigate the detection robustness bottleneck that the object detector discards a portion of its performance on the clean image while gaining a very limited robustness from adversarial training. Empirical analysis from the loss change and gradient interference indicate that the detection robustness bottleneck is mainly attributed to the conflict between the object detector in learning clean images and adversarial images. 
It is hard for object detectors to learn both images well, so it needs a learning trade-off between them. 

\textls[-10]{In terms of the detection robustness bottleneck on both clean images and adversarial images, we propose the RobustDet method based on adversarially-aware convolution.
RobustDet utilizes an Adversarial Image Discriminator (AID) to generate different weights to clean images and adversarial images, which guides the generation of adversarially-aware convolutional kernels to adaptively learn robust features. RobustDet also employs the Consistent Features with Reconstruction (CFR)
to make the features of clean and adversarial images in the same distribution and empower the model to reconstruct the adversarial image into a clean image. This can further enhance the detection robustness.
Besides, experimental results show that our method can effectively alleviate the detection robustness bottleneck. It is demonstrated that our method can significantly improve the robustness of the model without losing the performance on clean images.}

\vspace{-6pt}
\section*{Acknowledgement}
\vspace{-6pt}

This work was supported in part by NSFC (No.62006253, U21A20470, 61876224), National Key R\&D Program of China (2021ZD0111601).

\clearpage
\bibliographystyle{splncs04}
\bibliography{egbib}
\end{document}